\documentclass[letterpaper]{article} 
\usepackage{aaai23}  
\usepackage{times}  
\usepackage{helvet}  
\usepackage{courier}  
\usepackage[hyphens]{url}  
\usepackage{graphicx} 
\usepackage{natbib}  
\usepackage{caption} 
\usepackage{algorithm}
\usepackage{algorithmic}

\usepackage[linewidth=1pt]{mdframed}
\usepackage{times}
\usepackage{latexsym}
\usepackage{booktabs}
\usepackage{multirow}
\usepackage[utf8]{inputenc}
\usepackage{pifont}
\usepackage{soul}
\usepackage{comment}

\usepackage{bm}

\usepackage{graphics}
\usepackage{graphicx}
\usepackage{subfigure}
\usepackage{amsmath}
\usepackage{amssymb}
\usepackage{newfloat}
\usepackage{listings}
\DeclareCaptionStyle{ruled}{labelfont=normalfont,labelsep=colon,strut=off} 
\floatstyle{ruled}
\newfloat{listing}{tb}{lst}{}
\floatname{listing}{Listing}
\title{Generating Coherent Narratives by Learning Dynamic and Discrete Entity States with a Contrastive Framework}

\author{
Jian Guan$^1$, Zhenyu Yang$^2$, Rongsheng Zhang$^3$, Zhipeng Hu$^3$ and Minlie Huang$^1$\thanks{Corresponding author}
}
\affiliations{
    \textsuperscript{\rm 1}The CoAI group, DCST; \textsuperscript{\rm 1}Institute for Artificial Intelligence; \textsuperscript{\rm 1}State Key Lab of Intelligent Technology and Systems;\\
    \textsuperscript{\rm 1}Beijing National Research Center for Information Science and Technology; \textsuperscript{\rm 1}Tsinghua University, Beijing 100084, China.\\
    \textsuperscript{\rm 2}Guangdong OPPO Mobile Telecommunications Corp., Ltd.
    \textsuperscript{\rm 3}Fuxi AI Lab, NetEase Inc., Hangzhou, China.\\

    {{j-guan19@mails.tsinghua.edu.cn;}} {{yangzhenyu@oppo.com;}}\\ {{\{zhangrongsheng, zphu\}@corp.netease.com;}} {{aihuang@tsinghua.edu.cn}} \\
}

\usepackage{bibentry}

\begin{document}

\maketitle

\begin{abstract}
Despite advances in generating fluent texts, existing pretraining models tend to attach incoherent event sequences to involved entities 
when generating narratives such as stories and news. 
We conjecture that such issues result from representing entities as static embeddings of superficial words, while neglecting to model their ever-changing states, i.e., the information they carry, as the text unfolds.
Therefore, we extend the Transformer model to dynamically conduct entity state updates and sentence realization for narrative generation. 
We propose a contrastive framework to learn the state representations in a discrete space, and insert additional attention layers into the decoder to better exploit these states. 
Experiments on two narrative datasets show that our model can generate more coherent and diverse narratives than strong baselines with the guidance of meaningful entity states.
\end{abstract}

\section{Introduction}
Generating open-ended texts that maintain long-range coherence is important for myriad natural language generation applications such as narrative generation. 
The task requires, given a short input, creating a text with a sequence of coherent events involving several interleaved entities~(e.g., characters, organizations, locations, etc.
). As the text unfolds, the entities encounter different events, enrich their respective information, and accordingly change readers' expectations about them, thus playing a central role to make the text coherent
~\cite{grosz1995centering}. 
Arguably, the ability to dynamically predict unseen events attached to different entities  is indispensable for generation~\cite{henaff2017tracking} but has not yet been widely investigated.

\begin{table}[!t]
\scriptsize
    \centering
    \begin{tabular}{p{225pt}}
    \toprule
    \textbf{\textsf{Example 1:}} ... After \textbf{\textcolor{red}{Bobby} 
    is injured}, the scene flashes to \textbf{{\textcolor{red}{Bobby}’s funeral}}. \textbf{{\textcolor{red}{Bobby} then goes to a graveyard}} with the family and begins to tell of a woman that ... \\
    \midrule
    \textbf{\textsf{Example 2:}} ... The Tokyo District Court found \textbf{\textcolor{blue}{Samsung} guilty} last week for  ``intentionally causing serious injury to the plaintiffs by copying, copying, or causing injury to a trademarked trademark.'' The court ruled that \textbf{\textcolor{blue}{Samsung} violated the patent on all \textcolor{cyan}{Apple}'s mobile devices}, including the iPhone and iPad ... 
    \textbf{\textcolor{cyan}{Apple} has been fighting the ruling} since Wednesday ...\\ 
    \bottomrule
    \end{tabular}
    \caption{Two examples generated by BART fine-tuned on the Wikiplots and CNN News datasets, respectively, where conflicting events are attached to some entities~(in \textbf{bold}). Different entity mentions are marked in different colors.}
    \label{tab:exam}
\end{table}

Typical generative models such as BART~\cite{bart} are trained to learn co-occurrence between tokens, which are represented as learnable embeddings.
As exemplified in Table~\ref{tab:exam}, BART can easily capture dependencies between 
common words such as ``injured'' and ``funeral,'' 
but attaches incoherent event sequences to the involved entities. We conjecture that such issues arise from representing entities as no more than static embeddings of superficial words throughout whole texts, while neglecting to model the change of information these entities carry, i.e., their states~\cite{ji2017dynamic}. Since the same superficial words can co-occur with different events (and vice versa), it is necessary for generating coherent texts to model 
dependencies between events and specific states instead of word embeddings. For instance, in the first example of Table~\ref{tab:exam}, after ``Bobby's funeral'', the model should learn to update its estimate of Bobby's state from being \textit{alive} to \textit{dead} and then attach such events to him as ``lying in a coffin'' instead of ``going to a graveyard.'' Similarly, in the second example, Apple and Samsung transit to different states after ``the court rule that Samsung violated the patent on all Apple's...,'' 
and the model should attach the event ``fighting the ruling'' to Samsung instead of Apple.
This work proposes a generation model that incorporates dynamic entity states to improve coherence. 
We extend the Transformer decoder to update entity states after each sentence, which then serve to guide the subsequent decoding. We design a novel contrastive framework to learn the state representations. 
Instead of conditioning on continuous state representations for generation~\cite{clark2018neural}, we use a set of discrete state vectors to represent entity states, which are a natural fit for state transitions.  Discrete states also encourage effective use of the latent space, and alleviate excessive focus on local and imperceptible details.


The contrastive framework is designed 
to pull the state of an entity 
close to events that can be attached to the entity in the representation space. To abstract high-level event features, we adopt an external encoder to encode each sentence in a mini-batch to obtain the corresponding 
entity-aware
event representations for different entities in the sentence. 
At the end of each sentence, we first learn to predict which entity to mention in the following sentence. 
We then learn the current state representation of the entity using a contrastive  objective by regarding the representation of the following event attached to it as the positive and all others in the same mini-batch as negatives. 
During inference, we look up the closest state vector to the state representation in the pre-defined discrete latent space, and feed the state vector
into the decoder along with the word embedding of the entity mention to guide the following sentence realization. The hidden states of the decoder get access to input entity states using not only the vanilla self-attention layer, but also a state attention layer inserted into each decoder block. The additional layer narrows the attention scope to only prefix entity states for explicitly modeling the dependencies between states and contextual events. 
In the training phase, we directly use the closest state vector to the following event representation as input to keep the parallel training efficiency.
Our contributions are as follows:

\noindent\textbf{I.} We propose a novel generation model that learns dynamic and discrete \underline{\textbf{e}}ntity state \underline{{\textbf{r}}}epresentat\underline{\textbf{i}}ons with a \underline{\textbf{c}}ontrastive framework~(\textsc{Eric}). We equip the decoder with an additional attention layer to apply entity states to guide the decoding. 

\noindent\textbf{II.} Extensive experiments on two datasets show that the contrastive framework learns a set of meaningful entity state vectors corresponding to different clusters of events that can be attached to an entity, thus enabling \textsc{Eric} to generate more coherent and informative texts 
than strong baselines\footnote{The codes are available at \url{https://github.com/thu-coai/ERIC}.}. 

\section{Related Work}
\paragraph{Narrative Generation} 
Recent studies presented a series of multi-step generation models for narrative generation, which first planned intermediate sketches like keywords~\cite{yao2019plan}, semantic role labeling tags~\cite{fan-etal-2019-strategies} and keyword distributions~\cite{kang-hovy-2020-plan}, and then generated whole texts conditioned on them. 
\citet{ji2021discodvt} learned a sequence of discrete latent codes to abstract high-level discourse structures. Each code corresponds to a fixed-length span, which does not always agree with real text structures and makes it hard to model specific semantic dependencies. 
Some studies tried to incorporate external knowledge or reasoning models to guide commonsense story generation~\cite{guan2020knowledge, DBLP:conf/emnlp/XuPSPFAC20,ammanabrolu2021automated}, which may 
lack generalization to other domains such as news.  
Another line improved coherence by learning high-level representations of prefix sentences~\cite{li2015hierarchical, guan-etal-2021-long}, which does not emphasize the central role of entities. 

Prior studies on state tracking commonly adopted an external memory for state reads and writes. \citet{ji2017dynamic} and \citet{clark2018neural} updated the state of an entity conditioned on previous hidden outputs when encountering its mention, which does not apply to the parallel architecture of Transformer for training. \citet{rashkin2020plotmachines} and \citet{papalampidi2022towards} performed state updates at the paragraph and chunk level, respectively, to alleviate but not eliminate the issue. In contrast, \textsc{Eric} learns entity states through entity-aware event representations derived from an external encoder during training, thus well adapting to popular pretraining models, 
and achieving more frequent state updates. Furthermore, our work presents the first study for learning discrete entity states. 

\paragraph{Hierarchical Transformers} It is necessary to capture the hierarchical structure of natural language texts~\cite{ribeiro-etal-2020-beyond}. Previous work employed hierarchical attention networks for document classification~\cite{yang-etal-2016-hierarchical} and machine translation~\cite{miculicich-etal-2018-document}. 
\citet{guo-etal-2019-hierarchical} adopted a hierarchy network to build document embeddings on top of sentence embeddings for document mining. Similarly, HIBERT~\cite{zhang2019hibert} incorporated a hierarchical architecture to BERT~\cite{devlin2018bert} for document summarization. 
These models aim to enhance encoders for modeling long inputs and are less adaptive for generation. 
Recent studies tried to shorten sequences in intermediate decoder blocks to learn high-level representations for text classification~\cite{DBLP:conf/nips/DaiLY020} and generation~\cite{nawrot2021hierarchical}. 
\citet{hu2022planet} incorporated dynamically planned bag-of-words to guide the text realization without considering dependencies between words. In comparison, the proposed framework enables the learned entity states to serve as a sentence-level guidance for generation. 

\paragraph{Contrastive Learning} Contrastive learning has become popular in unsupervised visual and textual representation learning~\cite{hadsell2006dimensionality}. 
The representations are learned by making two objects augmented from the same data point close in the vector space, and objects from others as distant as possible. 
CERT~\cite{fang2020cert} adopted an instance-level data augmentation technique~(i.e., back-translation). 
ConSERT~\cite{yan-etal-2021-consert} and SimCSE~\cite{gao-etal-2021-simcse} proposed  directly adding noise to inner representations of BERT to construct positive pairs, leading to better performance and higher computation efficiency. 
In this paper, we develop a novel contrastive framework to learn discrete entity states for text generation.

\begin{figure*}[t]
  \centering
\includegraphics[width=0.9\linewidth]{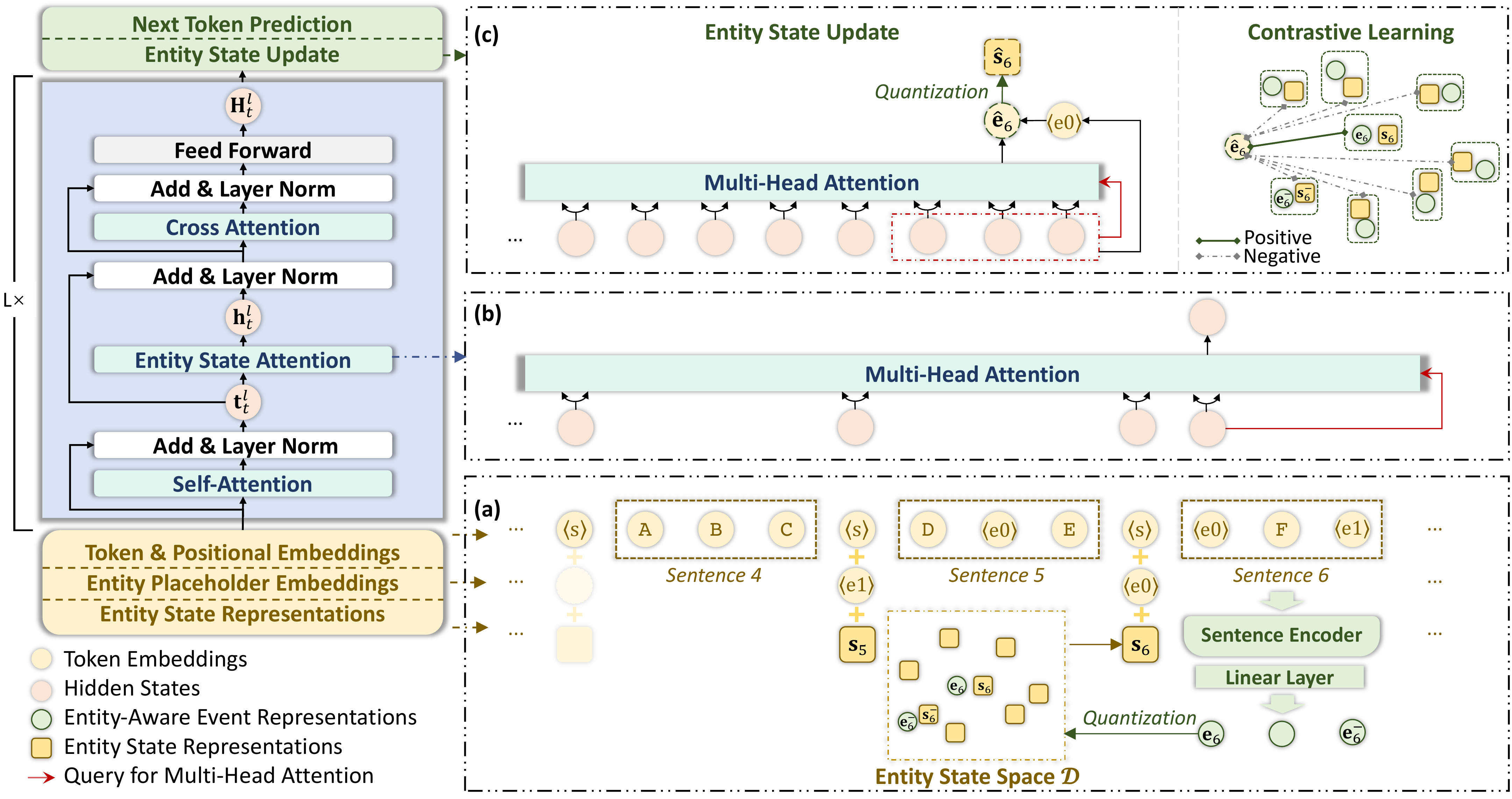}
  \caption{Model overview of the \textsc{{Eric}} decoder, which \textbf{(a)} adds a special token $\rm\langle s\rangle$ before each sentence for incorporating state representations of the following mentioned entities, \textbf{(b)} inserts an entity state attention layer into each Transformer block that narrows the attention scope to only the prefix entity states, and \textbf{(c)} learns an entity state update module with a contrastive objective. 
  We omit the input encoder of \textsc{Eric} for simplicity. }
  
  \label{fig:model}
\end{figure*}

\section{Methodology}
\subsection{Task Definition and Model Overview}
Our task is as follows: given a short input such as a beginning $X=(x_1,x_2,\cdots,x_M)$, the model should generate a coherent multi-sentence text $Y=(y_1,y_2,\cdots,y_L)$~(each $x_i$ or $y_i$ is a token). We notice that entity words (e.g., character names) often consist of rare tokens, and the same entity may have different mentions~(e.g., ``Bruce Wayne,'' ``Bruce'' and ``Wayne''). To better track entity states, we use a two-stage generation framework~\cite{hermann2015teaching} that first generates a coarse text with all mentions of each entity replaced by a unique placeholder such as ``$\langle\rm e0\rangle, \langle\rm e1\rangle,\cdots$,'' and then generates the mention for each one. 
We denote the coarse version of the target as $Y^e=(y^e_1, y^e_2,\cdots,y_T^e)$.  




To generate $Y^e$ from $X$, generative models such as BART are commonly optimized to 
minimize the negative log-likelihood $\mathcal{L}_{\text{LM}}$ of $Y^e$ as follows: 
\begin{align}
    \mathcal{L}_{\text{LM}}&= -\sum_{t=1}^T\log {P}({y}^e_t|{y}^e_{<t}, X).
\end{align} 
Further, \textsc{Eric} conducts entity state updates and sentence realization in an auto-regressive manner. At the end of each sentence, we incorporate the state vector of the entity mentioned in the following sentence to the decoder~($\S$~\ref{inco}), which then guides the subsequent generation with an entity state attention layer inserted into each decoder block~($\S$~\ref{stateattn}). We design a contrastive objective for learning to predict the state of the following mentioned entity conditioned on prefix decoder outputs~($\S$~\ref{cllearn}).
Although we only consider sentences as segments in this paper, our algorithm can easily be  extended to other syntactic levels such as paraphrases. Figure~\ref{fig:model} shows the model overview of the first training stage. 
In the second stage, we train another standard Transformer model to generate mentions for anonymous entities~($\S$~\ref{second}).


\subsection{Incorporating Entity States}\label{inco}

Suppose that ${Y^e}$ consists of $N$ sentences, denoted from ${Y}^e_1$ to ${Y}^e_N$~(e.g., \texttt{ABC} in Figure~\ref{fig:model}). We insert a special token $\langle \texttt{s}\rangle$ at the beginning of each sentence ${Y}^e_n$~($n=1,2\cdots, N$)~\cite{guan-etal-2021-long}, which is designed to incorporate entity state representations to guide the realization of ${Y}^e_n$. 
Formally, we set the decoder input $\textbf{H}^0_t$ as follows:
\begin{align}
    \textbf{H}^0_t&=\left\{
\begin{matrix}
\textbf{E}_t+\textbf{E}(p_n)+\textbf{s}_n, \hfill&\text{if}~y^e_t = \langle \texttt{s}\rangle|_n\hfill\label{input_eq}\\
\textbf{E}_t,\hfill &\text{otherwise}
\hfill\end{matrix}
\right.\\
\textbf{E}_t&=\textbf{E}(y^e_t)+\textbf{P}(y^e_t),
\end{align}
where 
$\textbf{E}_t$ is the sum of the token and positional embeddings of the $t$-th token $y^e_t$, $\langle \texttt{s}\rangle|_n$ means the $n$-th special token, $p_n$ is the placeholder token of the entity mentioned in $Y_n^e$, $\textbf{E}(p_n)$ and $\textbf{s}_n$ are the token embedding and state vector of $p_n$, respectively. When $Y_n$ does not mention any entities~(e.g., sentence 4 in Figure~\ref{fig:model}), we set both $\textbf{E}(p_n)$ and $\textbf{s}_n$ to zero vectors. While $Y_n$ contains multiple placeholders~(e.g., sentence 6 in Figure~\ref{fig:model}), we randomly sample one as $p_n$.  
Our algorithm also adapts to multiple entities, which is left to future work. 



We pre-define a discrete entity state space $\mathcal{D}\in\mathbb{R}^{K\times D}$, 
which consists of $K$ entity states with each state represented as a normalized $D$-dimensional vector $\textbf{d}_i$~($i=1,2\cdots, K$). As shown in Figure~\ref{fig:model}, we pass a sentence $Y^e_n$ through an external bidirectional sentence encoder, and map the hidden output at the position of $p_n$ ~(e.g., $\langle\rm e0\rangle$ in Figure~\ref{fig:model}) 
to $D$ dimensions using a linear layer and normalize it as the entity-aware event representation, denoted as $\textbf{e}_n$. We then derive $\textbf{s}_n$, the state vector of $p_n$, through the nearest neighbour look-up from $\mathcal{D}$: 
\begin{align}
\textbf{s}_n&=\textbf{d}_k, \text{where}~k=\mathop{\text{argmin}}\limits_{\textbf{d}_j\in\mathcal{D}} \textbf{d}_j \cdot \textbf{e}_n,
\end{align}
where we have normalized both $\textbf{d}_j$ and $\textbf{e}_n$  onto the unit hyper-sphere. We use the straight-through trick~\cite{bengio2013estimating} to allow the gradient to back-propagate to the sentence encoder, i.e., directly copying the gradient of $\textbf{s}_n$ to $\textbf{e}_n$~\cite{van2017neural}.

\subsection{Entity State Attention Layer}\label{stateattn}

In order to explicitly exert entity states on text  generation,  
we insert an additional entity state attention layer between the vanilla self-attention layer and cross-attention layer in each decoder block. Assuming that the decoder consists of $L$ blocks, let $\textbf{t}_{t}^{l}$ denote the layer-normalized output of the self-attention layer in the $l$-th block at the $t$-th position~$(t=1,2,\cdots, T,~ l=1,2,\cdots, L)$. The entity state attention layer allows $\textbf{t}_{t}^{l}$ to attend to only those hidden states corresponding to prefix entity states:
\begin{align}
\textbf{h}_t^l = \text{A}(\text{Q}=\textbf{t}_t^l, \text{K}/\text{V}=\{\textbf{t}_{t'\leqslant t}^l|y_{t'}^e=\langle \texttt{s}\rangle\}),
\end{align}
where $\textbf{h}_t^l$ is the output hidden state of the state attention layer, $\text{A}(\cdot)$ means the multi-head attention mechanism~\cite{vaswani2017attention}, Q, K and V are the corresponding query, key and value vectors. 
In this way, \textsc{Eric} predicts the next tokens with the guidance of entity states using an individual attention network besides the standard self-attention, enhancing its ability to model dependencies between sentence realization and entity states.

\subsection{Entity States Learning}\label{cllearn}
\textsc{Eric} uses entity states derived from the following golden sentences for training. We add an entity state update module on top of the decoder to learn to predict entity states based on prefix information using a contrastive framework, which will be used to guide generation in the inference stage.

The entity state update module consists of two key components, which are used to predict the following mentioned entity $\hat{p}_n$ and its state ${\hat{\textbf{s}}}_n$~($n=1,2,\cdots, N$), respectively.
We adopt a linear layer to predict  the distribution of $\hat{p}_n$ over the vocabulary of all placeholders conditioned on the prefix, and minimize the prediction loss $\mathcal{L}_{\rm Ent}$ as follows:
\begin{align}
    \mathcal{L}_{\rm Ent}&=-\sum_{n=1}^N\text{log}P(\hat{p}_n=p_n),\\
    P(\hat{p}_n) &= \text{softmax}(\boldsymbol{W}_p{\textbf{q}}_{n-1}+\boldsymbol{b}_p),\\
    \textbf{q}_{n-1}&=\text{MeanPool}(\{\textbf{H}_t^L\}_{n-1}),
\end{align}
where $\textbf{q}_{n-1}$ is a context summary vector derived by applying mean-pooling on $\{\textbf{H}_t^L\}_{n-1}$, i.e., the set of hidden outputs of the $(n-1)$-th sentence, 
$\boldsymbol{W}_p$ and $\boldsymbol{b}_p$ are trainable parameters. We add a special token $\langle\texttt{none}\rangle$ into the placeholder vocabulary with a constant zero embedding to indicate that no entities will be mentioned in the following sentence. We decide $\hat{p}_n$ by taking a sample from $P(\hat{p}_n)$, and predict its state vector $\hat{\textbf{s}}_n$ as follows: 
\begin{align}
    \hat{\textbf{s}}_n&=\textbf{d}_k, \text{where}~k=\mathop{\text{argmin}}\limits_{\textbf{d}_j\in\mathcal{D}}~\textbf{d}_j\cdot\hat{\textbf{e}}_n.\label{sss}\\
    \hat{\textbf{e}}_n
    &=\text{Normalize}\big(\text{A}(\text{Q}=\textbf{q}_{n-1}+\textbf{E}(\hat{p}_n),\\ &~~~~~\text{K}/\text{V}=\{\textbf{H}^L_{t}\}_{1:n-1})\big),\nonumber
\end{align}
where 
$\hat{\textbf{e}}_n$ is the continuous state representation of $\hat{p}_n$ before quantization, and $\{\textbf{H}_t^L\}_{1:n-1}$ is the set of hidden states of the first $n-1$ sentences.
To learn the state representation, we design a contrastive framework  
to draw $\hat{\textbf{e}}_{n}$ close to the following event representation $\textbf{e}_n$ and 
keep it away from others derived from different sentences or corresponding to different entities in the same mini-batch~(e.g., $\textbf{e}_6^-$ in Figure~\ref{fig:model}). 
To back-propagate gradients to learn the state space $\mathcal{D}$, we set each positive or negative to a joint representation that combines the event representation $\textbf{e}_n$ and its nearest state vector $\textbf{s}_n$ in $\mathcal{D}$. In this way,  they can be optimized in the same direction and forced to  distributed uniformly in $\mathcal{D}$, thus gaining better expressiveness.
We formulate the  contrastive objective $\mathcal{L}_{CL}$ as the following infoNCE loss~\cite{oord2018representation}:
\begin{align}
    \mathcal{L}_{\text{CL}}&=-\frac1{N}\sum_{n=1}^N\text{log}\frac{\text{exp}({\hat{\textbf{e}}_n\cdot\textbf{c}_n/\tau)}}{\sum\limits_{{\textbf{c}}^*_n\in C}\text{exp}{(\hat{\textbf{e}}_n\cdot\textbf{c}_n^*/\tau)}},\\\label{loss_cl}
    \textbf{c}_n &= \text{Normalize}({\textbf{e}_n+\textbf{s}}_n),
\end{align}
where $\textbf{c}_n$ is the positive joint representation for $\hat{\textbf{e}}_n$, $C$ is the set of all joint representations in the mini-batch, and $\tau$ is the adjustable temperature. 

Once we obtain $\hat{\textbf{e}}_n$, the corresponding discrete state vector $\hat{\textbf{s}}_n$ can be obtained by quantizing $\hat{\textbf{e}}_n$ to $\mathcal{D}$ using the nearest neighbour look-up as shown in Eq.~\ref{sss}. 
And $\hat{p}_n$ and $\hat{\textbf{s}}_n$ will be taken as input in Eq.~\ref{input_eq} to guide the generation in the inference stage.
In summary, we train the input encoder, the sentence encoder and the decoder jointly with the following overall loss function:
\begin{align}
    \mathcal{L} = \mathcal{L}_{\text{LM}} + \lambda_1\mathcal{L}_{\text{Ent}} + \lambda_2\mathcal{L}_{\text{CL}},\label{loss}
\end{align}
where $\lambda_1$ and $\lambda_2$ are adjustable sale factors.

\subsection{Entity Mention Generation}\label{second}
This stage requires generating superficial entity mentions for placeholder tokens in $Y^e$. 
We only use a standard Transformer model for this stage since it is not the main focus of this work. Other training techniques can be also easily applied~\cite{fan-etal-2019-strategies}.  
Formally, given $X$ and $Y^e$, the model should generate a sequence of pairs of a placeholder token followed by its corresponding superficial word in the order that they are mentioned. For example, for the text \textit{``$\langle e\textit{0}\rangle$ and his girlfriend $\langle e\textit{1}\rangle$ take a romantic vacation to a cabin. While in the cabin, $\langle e\textit{0}\rangle \cdots$,''} the golden output is \textit{``$\langle e\textit{0}\rangle$Ash Williams$\langle e\textit{1}\rangle$Linda$\langle e\textit{0}\rangle$Ash $\cdots$.''} After obtaining the entity
mentions, we can insert them back into $Y^e$ to get the whole text $Y$. 

\section{Experiments}
\subsection{Datasets} 
We evaluate \textsc{Eric} on two English narrative datasets, 
Wikiplots\footnote{\url{www.github.com/markriedl/Wikiplots}} and CNN News~\cite{hermann2015teaching}. 
Wikiplots collected story plots from Wikipedia. 
We use the official split of Wikiplots. 
We follow \citet{ji2021discodvt} to split each sentence into sequential elementary discourse units and regard them as sentences for experiments on Wikiplots.
CNN News consists of online newspaper articles collected from CNN. We process the dataset using the script provided by~\citet{tan2021progressive} and split the dataset randomly by 18:1:1 for training/validation/testing.
For both datasets, we take the first sentence as input to generate the rest. 
We remove those texts whose outputs contain less than five sentences, and truncate each output to at most fifteen sentences. 
We use spaCy\footnote{\url{https://spacy.io/usage/linguistic-features\#named-entities}} to identify people and organization names in a text as entity mentions. 
If the string of an entity mention is included in another, we replace them with the same placeholder. There are about 98\% of examples and 50\% of sentences that mention at least one entity on both datasets. More statistics are shown in Table~\ref{stat}, where we count tokens using the NLTK tokenizer~\cite{loper2002nltk}. More details are shown in Appendix~\ref{datapro}.

\begin{table}[!ht]
\scriptsize
\centering
\begin{tabular}{l|ccc|ccc}
\toprule

\multirow{2}{*}{\textbf{Datasets}}&\multicolumn{3}{c|}{\textbf{Wikiplots}}&\multicolumn{3}{c}{\textbf{CNN News}}\\
&\textbf{Train}&\textbf{Val}&\textbf{Test}&\textbf{Train}&\textbf{Val}&\textbf{Test}\\
\midrule
\textbf{\#Example}&76,826&4,327&4,324&82,836&4,602&4,602\\
\midrule
\textbf{Ipt Len.}&24.6&24.6&24.6&32.9&33.3&33.0\\
\midrule
\textbf{Opt Len.}&237.1&236.5&237.1&346.0&344.9&344.5\\
\textbf{Avg. \#Sen.}&12.6&12.7&12.6&14.3&14.2&14.3\\
\textbf{Avg. \#Ent.}&6.8&6.7&6.7&7.7&7.6&7.7\\
\bottomrule
\end{tabular}
\caption{Example numbers, average lengths~(\textit{Len.}) of inputs~(\textit{ipt}) and {outputs}~(\textit{opt}), average numbers of sentences~(\textit{Sen.}) and distinct entities~(\textit{Ent.}) in outputs 
for \textit{{train}}ing, \textit{{val}}idation and \textit{test}ing, respectively.}
\label{stat}
\end{table}

\subsection{Implementation}
Our algorithm adapts to all generative models with auto-regressive decoders.
Due to limited computational resources, we use BART$_{\rm Base}$'s pretrained checkpoint for initialization. The sentence encoder is initialized using the pretrained parameters of the BART$_{\rm Base}$ encoder. We set the number of discrete entity states in $\mathcal{D}$ to 512, the dimension of state vectors to 128, the maximum number of distinct entities in a text to 100, $\tau$ in Eq.~\ref{loss_cl} to 0.1, and $\lambda_1$/$\lambda_2$ in Eq.~\ref{loss} to 1/1. These settings lead to 3\% more parameters of \textsc{Eric} than BART$_{\rm Base}$\footnote{We do not count the parameters of the sentence encoder since it is not used during inference.}. 
For both stages, we set the batch size to 12, the maximum sequence length to 512, and the learning rate to 1e-4.
We decide the hyper-parameters based on the performance on the validation set.

During inference, we use top-$p$ sampling~\cite{holtzman2019curious} with $p=0.9$ for both generation stages. In the second stage, when the model fails to generate a mention word for a certain placeholder, we complete the output of the first stage using the last word generated for this placeholder if it has been mentioned before~(about 2.5\% of cases), or a random name otherwise~(about 1.1\% of cases). 

\subsection{Baselines}
We compared \textsc{Eric} with the following models: \textbf{(1) Seq2Seq:} It has the same architecture as BART$_{\rm Base}$ but is initialized randomly. 
\textbf{(2) BART:} It is fine-tuned on the downstream datasets with the standard language modeling objective.
\textbf{(3) PlanAhead:} It first plans a keyword distribution and then combines the planned distribution with the language model prediction using a gated mechanism~\cite{kang-hovy-2020-plan}. We use BART$_{\rm Base}$ as the backbone model and add additional parameters for planning and distribution combination. 
\textbf{(4) \textsc{Hint}:} It incorporates high-level sentence representations into BART$_{\rm Base}$ 
and uses sentence similarity prediction and sentence order discrimination to learn these representations~\cite{guan-etal-2021-long}. 
\textbf{(5) \textsc{DiscoDVT}:} It extends BART$_{\rm Base}$ to represent the discourse structure using a sequence of latent codes with learnable embeddings
~\cite{ji2021discodvt}. 
We do not limit the minimum length of the latent code sequence like in the original paper.
\textbf{(6) SimCTG:} It adds 
a contrastive objective which tries to distribute the hidden outputs of BART$_{\rm Base}$  uniformly in the representation space~\cite{su2022contrastive}. 


Besides the above baselines, we conduct ablation tests on Wikiplots by removing the proposed components respectively. 
For fair comparison, we insert the special token $\langle \rm \texttt{s}\rangle$ before each sentence for all baselines.  
During evaluation, we remove all special tokens from the generated texts. 


\begin{table*}[!ht]
\scriptsize
\centering
\begin{tabular}{l||cc|cc|c||ccc||cc||c||c}
    \toprule
    \textbf{Models}&\textbf{B-1$\uparrow$}&\textbf{B-2$\uparrow$}&\textbf{MSJ-1$\uparrow$}&\textbf{MSJ-2$\uparrow$}&\textbf{{MAUVE$\uparrow$}}&\textbf{Rpt-16$\downarrow$}&\textbf{Rpt-32$\downarrow$}&\textbf{Rpt-64$\downarrow$}&\textbf{D-3$\uparrow$}&\textbf{D-4$\uparrow$}&\textbf{Zipf}&\textbf{Len}\\
    \midrule
    \midrule
    \multicolumn{13}{c}{\textbf{Dataset: Wikiplots}}\\
    \midrule
    \textbf{Seq2Seq}&26.33&10.28&56.87&39.94&74.56&20.02&32.99&41.61&68.30&89.79&1.26&199\\
    \textbf{BART}&28.33&11.66&58.96&41.16&76.88&17.82&30.74&39.47&70.26&90.51&1.19&201\\
    \textbf{PlanAhead}&27.52&11.32&58.86&41.31&75.98&17.40&30.27&38.90&71.84&91.27&1.17&193\\
    \textbf{\textsc{Hint}}&29.81&{12.22}&61.17&{42.33}&80.16&17.20&30.12&39.00&71.73&91.12&1.14&209\\
    \textbf{\textsc{DiscoDVT}}&28.76&11.76&60.89&42.29&78.01&17.07&29.81&38.56&73.22&91.76&1.13&204\\
    \textbf{SimCTG}&29.08&11.90&59.81&41.53&77.15&17.68&30.59&39.40&70.44&90.52&1.18&206\\    
    \midrule
    \textbf{\textsc{Eric}}&\textbf{31.81}&\textbf{12.90}&\textbf{63.93}&\textbf{42.94}&\textbf{87.88}&\textbf{16.22}&\textbf{28.12}&\textbf{37.06}&\textbf{75.94}&\textbf{93.02}&\textbf{1.11}&236\\
    \textbf{~~w/o StateAttn}&\underline{30.12}&\underline{12.26}&\underline{62.81}&\underline{42.84}&\underline{83.29}&16.75&\underline{28.56}&\underline{37.42}&\underline{75.41}&\underline{92.85}&\underline{1.12}&221\\   
    \textbf{~~w/o $\textbf{s}_n$}&28.84&11.70&59.16&40.84&83.18&16.62&28.90&37.43&72.42&91.36&1.15&206\\
    \textbf{~~w/o} $\textbf{s}_n$\& $p_n$&28.63&11.59&58.74&40.43&73.01&\underline{16.34}&28.86&37.93&72.97&91.54&\underline{1.12}&203\\
    \midrule
    \textbf{\textit{Truth}}&\textit{N/A}&\textit{N/A}&\textit{N/A}&\textit{N/A}&\textit{N/A}&\textit{13.63}&\textit{25.30}&\textit{34.75}&\textit{85.80}&\textit{97.10}&\textit{0.96}&\textit{266}\\
    \midrule
    \midrule
    \multicolumn{13}{c}{\textbf{Dataset: CNN News}}\\   
    \midrule
    \textbf{Seq2Seq}&31.99&14.42&68.25&47.68&85.20&16.06&27.40&36.88&74.02&91.42&1.11&271\\
    \textbf{BART}&32.18&14.66&68.20&47.37&\underline{87.25}&14.63&25.74&35.34&77.05&92.64&1.08&267\\
    \textbf{PlanAhead}&32.23&14.64&67.05&47.11&28.95&14.55&25.70&35.48&76.53&92.12&1.08&259\\
    \textbf{\textsc{Hint}}&\underline{33.07}&\underline{15.18}&\underline{69.29}&{47.74}&86.15&14.57&25.69&35.19&77.16&92.61&\underline{1.07}&275\\
    \textbf{\textsc{DiscoDVT}}&32.53&14.92&68.63&47.64&83.43&\underline{14.37}&\underline{25.50}&\underline{35.09}&\underline{77.47}&\underline{92.75}&\underline{1.07}&267\\
    \textbf{SimCTG}&32.50&14.79&68.91&\underline{47.75}&86.65&\underline{14.37}&25.56&35.11&76.92&92.48&\underline{1.07}&272\\    
    \midrule
    \textbf{\textsc{Eric}}&\textbf{33.83}&\textbf{15.28}&\textbf{70.47}&\textbf{48.09}&\textbf{91.02}&\textbf{14.29}&\textbf{25.16}&\textbf{34.81}&\textbf{78.16}&\textbf{93.01}&\textbf{1.06}&282\\
    \midrule
    \textbf{\textit{Truth}}&\textit{N/A}&\textit{N/A}&\textit{N/A}&\textit{N/A}&\textit{N/A}&\textit{12.28}&\textit{22.39}&\textit{32.22}&\textit{82.89}&\textit{94.61}&\textit{1.00}&\textit{341}\\
    \bottomrule
\end{tabular}
\caption{Automatic evaluation results. 
$\downarrow$ / $\uparrow$ means the lower/higher the better. The best performance is highlighted in \textbf{bold} the second is \underline{underlined}. {\textsc{Eric} w/o StateAttn} means removing the entity state attention layer. {\textsc{Eric} w/o $\textbf{s}_n$} means removing entity state representations in the decoder input along with the contrastive learning framework. {\textsc{Eric} w/o $\textbf{s}_n$ \& $p_n$} means further remove the next mentioned entity prediction module.} 
\label{auto-eva}
\end{table*}

\subsection{Automatic Evaluation}
\paragraph{Evaluation Metrics} We do not use perplexity for evaluation since the two-stage generation paradigm of \textsc{Eric} makes it intractable to assess the text probability. We use the following automatic metrics: 
\textbf{(1) BLEU~(B-n)}: It evaluates the $n$-gram overlap between generated and human-written texts~\cite{papineni2002bleu}, $n=1,2$. 
\textbf{(2) MS-Jaccard~(MSJ-n)}: It measures the similarity between two $n$-gram distributions of generated and human-written texts using the Jaccard Index between two multi-sets of $n$-grams~\cite{alihosseini2019jointly}, $n=1,2$.
\textbf{(3) MAUVE}: It measures the similarity between two text distributions of generated and human-written texts~\cite{mauve}, where text representations are derived from GPT2$_{\rm Base}$.
\textbf{(4) Token Repetition~(Rpt-n)}: It measures the repetition of generated texts by calculating the fraction of the identical token that occurs in the previous $n$ tokens~\cite{Welleck2020Neural}, $n=16,32,64$.
\textbf{(5) Distinct~(D-n)}: It measures the generation diversity using the ratio of distinct $n$-grams to all generated $n$-grams~\cite{DBLP:conf/naacl/LiGBGD16}, $n=3,4$. \textbf{(6) Zipf Coefficient~(Zipf):} It computes the unigram rank-frequency scale factor in generated texts~\cite{holtzman2019curious}. A value closer to 1 indicates that the generated texts are closer to real-world texts in unigram distribution.
Moreover, we also report the average number of generated tokens, denoted as \textbf{Len}.

\paragraph{Results}
Table~\ref{auto-eva} shows the results on 1000 randomly sampled generated examples. 
The higher BLEU scores of \textsc{Eric} indicate that it 
can generate more $n$-gram overlaps with reference texts than baselines. On the whole, the generation distribution of \textsc{Eric} is more similar to the ground truth in terms of both $n$-grams and machine-derived text representations, as shown by the higher MSJ and MAUVE scores.
Furthermore, the texts generated by \textsc{Eric} suffer from less repetition with lower token repetition ratios in various ranges and have better diversity. 
\textsc{Eric} also achieves better modeling of long-tail tokens~(e.g., entity mentions), with a Zipf score closer to 1. The superiority of \textsc{Eric} on both datasets proves its generalization for text generation with different lengths and domains. Additionally, we observe that the Wikiplots dataset has more long-tail tokens than CNN News with a lower Zipf score, which may account for the higher superiority of \textsc{Eric} on Wikiplots since more low-frequency entity mentions may make it harder for baselines to model the entity coherence implicitly.



For ablation tests, the entity state attention layer helps the decoder exploit entity state representations better, thereby improving performance on all metrics and particularly reducing short-range repetition. When removing $\textbf{s}_n$ or both $\textbf{s}_n$ and ${p}_n$, the BLEU and MSJ scores drop to the level of BART, suggesting the importance of tracking entity states. We also notice that they still have surprisingly less repetition than all baselines. Manual inspection finds that they tend to generate fewer coordinate combinations of identical entity names. For example, there are 14.8\% and 13.0\% of texts generated by BART and \textsc{Eric} w/o $\textbf{s}_n \& p_n$, respectively, that contain strings of the form \textit{``W and W''} (\textit{``W''} is a unigram). The phenomenon shows that the two-stage generation paradigm may make it easier to learn entity mention patterns. Significantly, \textsc{Eric} further surpasses the two ablation models with less repetition thanks to 
the modeling of entity state transitions which integrates high-level dependencies between events attached to involved entities.

\begin{figure}[!ht]
  \centering
\includegraphics[width=\linewidth]{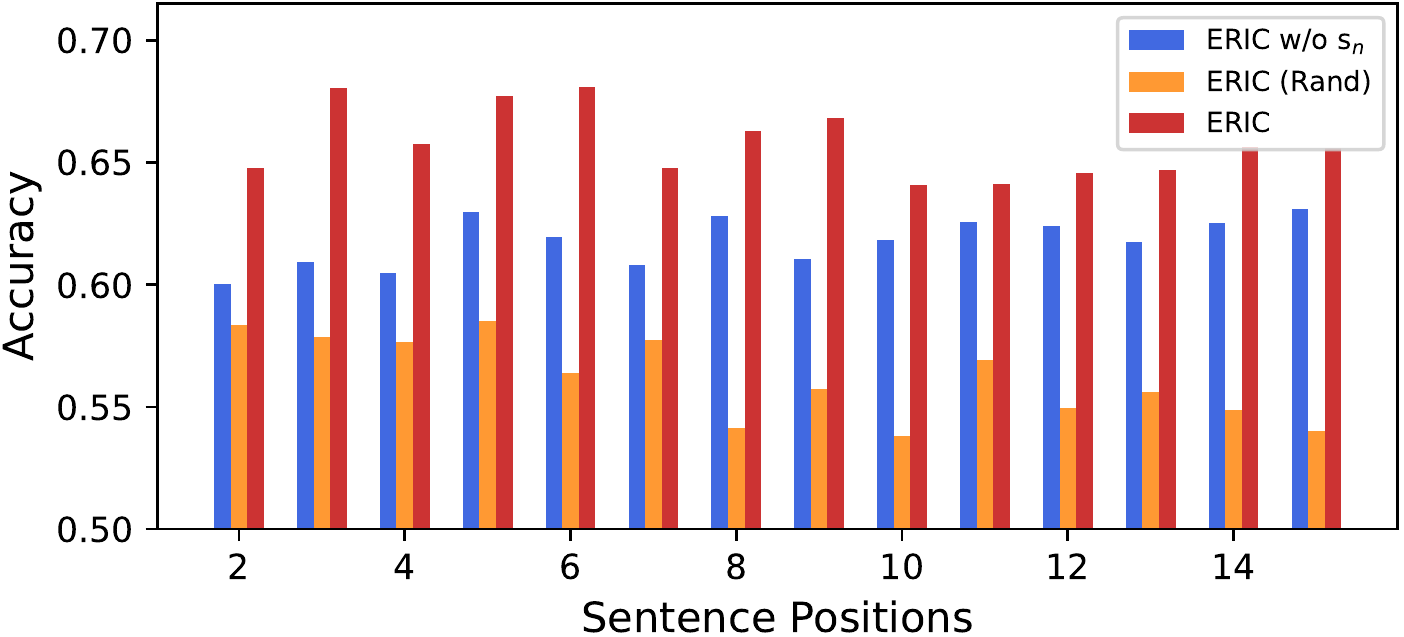}
  \caption{Accuracy of different models varying with positions of perturbed sentences on Wikiplots.} 
  \label{fig:acc}
\end{figure}

\paragraph{Entity Coherence Modeling} It is necessary to investigate whether tracking entity states helps better capture the entity coherence. To this end, on the test set of Wikiplots with masked entity mentions, we replace the first entity placeholder in each sentence in order with another that has been mentioned before randomly~(with the prefix not perturbed). Figure~\ref{fig:acc} plots the accuracy that a model gives a lower probability to the perturbed sentence than the original one along with the prefix. We calculate text probabilities using the following mentioned entities~(i.e., $p_n$) as input for all models, and using the entity states predicted by the model~(i.e., $\hat{\textbf{s}}_n$) as input for \textsc{Eric} and randomly sampled entity states for \textsc{Eric}~(Rand). We use the models after the first-stage training for this experiment.
\begin{figure*}[!ht]
  \centering
\includegraphics[width=\linewidth]{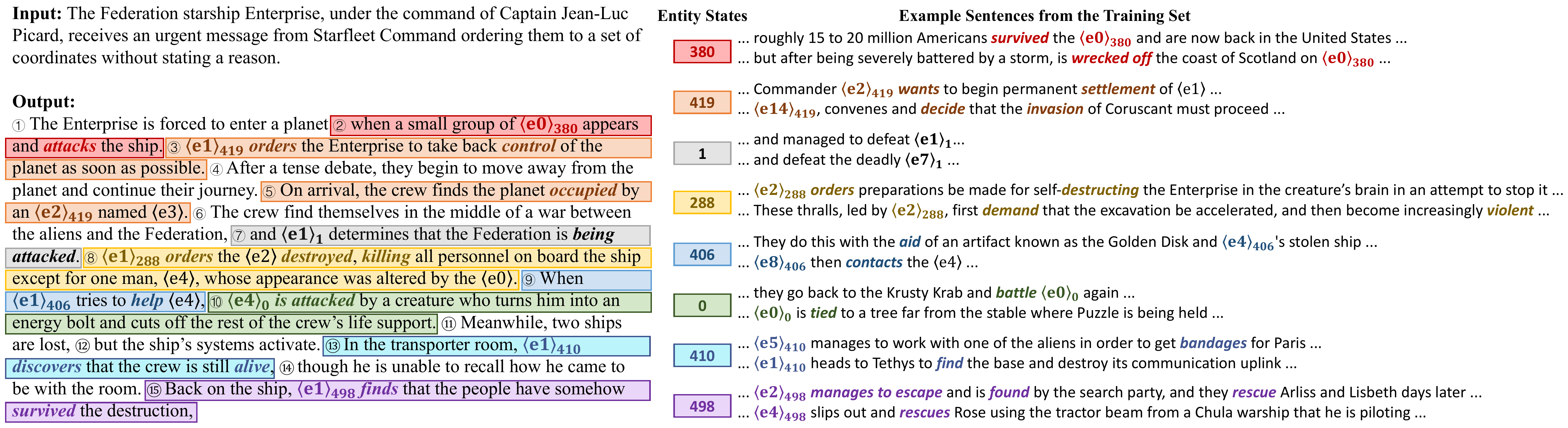}
  \caption{A case generated by \textsc{Eric} on Wikiplots~(\textbf{Left}) and example sentences from the training set where some entity has the corresponding state~(\textbf{Right}).
  The subscript under each entity placeholder token~(e.g., $\langle\rm {e0}\rangle_{\rm 380}$) denotes its state ID. 
  We highlight semantically correlated keywords between the generated case and example sentences in \textit{\textbf{italic}} type. For each entity state, two example sentences are manually selected from top ten sentences with the closest event representations to the state vector.}
  \label{fig:case}
\end{figure*}

We observe that \textsc{Eric} outperforms \textsc{Eric} w/o $\textbf{s}_n$ significantly~($p<0.01$), especially for the first several sentences. And its performance drops substantially when using random entity states, suggesting that \textsc{Eric} has a better ability to model entity coherence with the guidance of meaningful entity states. For example, when a placeholder $\langle\rm e1\rangle$ is replaced to $\langle\rm e0\rangle$, the event that is attached to $\langle\rm e1\rangle$ originally may disagree with the state of $\langle\rm e0\rangle$. In this case, it is easier for \textsc{Eric} to capture such incoherence issues by tracking their internal states. Appendix~\ref{state_ana} shows more analysis. 

\subsection{Manual Evaluation}
We conduct pair-wise comparisons between \textsc{Eric} and three strong baselines including BART, \textsc{Hint} and \textsc{DiscoDVT}, 
with each pair of models compared conditioned on 200 inputs randomly sampled from the test set of Wikiplots. We hired workers on Amazon Mechanical Turk~(AMT) to give a preference (win, lose, or tie) in terms of informativeness~(interesting, diverse and rich details) and coherence~(reasonable inter-sentence dependencies, e.g., causal and temporal relationships). Each pair is annotated by three workers independently. Workers are also asked to annotate whether each text exhibits entity issues such as attaching conflicting or repetitive events to an entity.
We use majority voting to make final decisions among the workers. 
Appendix \ref{instruction} shows the annotation instruction.

\begin{table}[!h]
\scriptsize
\centering
\begin{tabular}{l|cc|c}
    \toprule
    \multirow{2}{*}{\textbf{\textsc{Eric} vs.}}&{\textbf{{Informativeness}}}&{\textbf{Coherence}}&{\textbf{Entity Issues}}\\
    &\textbf{\textbf{Win}}~/~\textbf{\textbf{Lose}}~/~\textbf{\textbf{Tie}}&\textbf{\textbf{Win}}~/~\textbf{\textbf{Lose}}~/~\textbf{\textbf{Tie}}&\textbf{\textsc{Eric}}~/~\textbf{the Other}\\
    \midrule
    \midrule
    \textbf{BART}&$55^{**}_{0.50}~/~33~/~12$&$57^{**}_{0.42}~/~35~/~8$&$61^{**}_{0.45}/~87_{0.47}$ \\
    \midrule
    \textbf{\textsc{Hint}}&$58^{**}_{0.44}~/~31~/~11$&$56^{**}_{0.47}~/~37~/~8$&$68^{*}_{0.43}/~81_{0.35}$\\
    \midrule
    \textbf{\textsc{DiscoDVT}}&$57^{**}_{0.47}/~32~/~11$&$54^{**}_{0.42}~/~34~/~12$&$56^{**}_{0.48}/~83_{0.52}$\\
    \bottomrule    
\end{tabular}
\caption{Manual evaluation results on Wikiplots. \textbf{Left:} Percentages~(\%) of \textit{win}, \textit{lose} or \textit{tie} when comparing \textsc{Eric} with a baseline. \textbf{Right:} Percentages~(\%) of texts that suffer from entity issues. The subscripts are Fleiss’ kappa~\cite{Fleiss1971Measuring}. The mark $^{*}$/$^{**}$ means \textsc{Eric} outperforms the baseline significantly with $p<0.05/0.01$~(sign test).}
\label{man_eva}
\end{table}

As shown in Table~\ref{man_eva}, all results show fair or better inter-annotator agreement ($\kappa\geqslant0.2$). 
\textsc{Eric} can generate more abundant details with  more coherent plots significantly~($p<0.01$), and suffers from about 20\% fewer entity issues than baselines by tracking entity states explicitly.




\subsection{Case Study}
Figure~\ref{fig:case} shows a case to investigate the correspondence between entity states and texts. We conclude that: \textbf{(1) The entity states learn meaningful correspondence to specific events.} As shown in the right part of Figure~\ref{fig:case}, {\textit{state 419}} relates to ``occupying some regions,'' {\textit{state 380}} means the entity may be an ``attacker'' while \textit{state 1} stands for ``being defeated.'' \textbf{(2) The entity states effectively guide text generation.} For instance, $\langle\rm e0\rangle_{380}$ in the second sentence ``attacks'' the ship. Both $\langle\rm e1\rangle_{419}$ in the third sentence and $\langle\rm e2\rangle_{419}$ in the fifth sentence intend to ``occupy'' the planet. \textbf{(3) Tracking entity states helps maintain long-range coherence for text generation.} For example, the protagonist $\langle\rm e1\rangle$ behaves with reasonable state transitions throughout the text, forming a coherent plot. Appendix~\ref{case_study} shows the generation results of baselines.

\section{Conclusion}
We present the first study to track entity states in large pretraining models for narrative generation. The proposed model \textsc{Eric} dynamically updates entity states at the sentence level, where each state associates with a cluster of events that can be attached to the entity. Then these states serve to guide the subsequent generation with an additional entity state attention layer. We design a contrastive framework to  learn entity-aware event representations and discrete state vectors jointly. 
\textsc{Eric} surpasses strong baselines in automatic and manual evaluation for story and news generation. Further analysis shows that \textsc{Eric} can better capture entity coherence and generate more coherent and informative texts with the guidance of meaningful state representations.

\section{Acknowledgements}
This work is supported by the Key Research and Development Program of Zhejiang Province (No. 2022C01011). This work was supported by the National Science Foundation for Distinguished Young Scholars (with No. 62125604) and the NSFC projects (Key project with No. 61936010 and regular project with No. 61876096). This work was also supported by the Guoqiang Institute of Tsinghua University, with Grant No. 2019GQG1 and 2020GQG0005. This work was also sponsored by Tsinghua-Toyota Joint Research Fund.



\bibliography{aaai23}
\clearpage
\appendix
    
\section{Experiment Settings}
\subsection{Data Processing}\label{datapro}

We conduct experiments on two existing public narrative datasets Wikiplots and CNN News, which are widely used for text generation. 
We notice that these datasets contain a few offensive plots, but we have not observed any personal information. We do not process these contents. We admit that there may still be unpredictable bias in these datasets. Many sentences in the Wikiplots stories contain multiple clauses, making it hard to derive meaningful event representations. Therefore, we use the dataset released by \citet{ji2021discodvt}, who split each sentence into sequential elementary discourse units~(EDUs, either a complete sentence or a parsed sub-sentence) for experiments on Wikiplots. 

On the other hand, we do not regard demonstrative pronouns~(e.g., ``I,'' ``her,'' ``their'') as entity mentions because we find that document-level coreference resolution is still difficult even for large pretraining models, 
and it will impair the model performance severely to learn entity state tracking using noisy entity labels. 
We believe that it will further boost the performance of \textsc{Eric} using more accurate coreference resolution models, which we leave for future work.

\subsection{Implementation}
We implement \textsc{Eric} using the same configuration as BART$_{\rm Base}$ provided by HuggingFace's Transformers\footnote{\url{https://github.com/huggingface/transformers}}. The vocabulary consists of 50,625 tokens. We regard $\langle\texttt{mask}\rangle$ in the original vocabulary as the sentence tokn $\langle\texttt{s}\rangle$.  It costs about 15 hours for training \textsc{Eric} on Wikiplots or CNN News. The results are based on one NVIDIA Tesla V100~(32GB memory) with a random single run.
\subsection{Baselines}
Our work is closely related to~\citet{ji2017dynamic,clark2018neural}. We do not compare against them because: (1) They are RNN-based models, which have significantly worse performance than current pretraining models. The comparison is unfair and meaningless. (2) They update the state of an entity conditioned on previous hidden outputs when encountering its mention, which does not apply to the parallel architecture of Transformer for training. Therefore, it is not easy to implement their algorithms based on Transformer-based pretraining models, which also motivates us to propose our model. 

On the other hand, we do not use \citet{rashkin2020plotmachines} or \citet{papalampidi2022towards} as baselines since they require text segments or entity names of ground-truth outputs as input to initialize the state representations. In contrast, \textsc{Eric} does not depend on specific inputs and is able to dynamically expand entities and perform state updates of these entities more frequently during the decoding process.


\section{Results on Validation Set}
Besides the performance on the test set reported in the main paper, we also provide the performance on the validation in Table~\ref{tab:val_res} for \textsc{Eric} and several strong baselines.

\begin{table}[!t]
\scriptsize
    \centering
    \begin{tabular}{l|ccccc}
    \toprule
    \textbf{Models}&\textbf{B-1}&\textbf{B-2}&\textbf{MAUVE}&\textbf{D-3}&\textbf{D-4}\\
    \midrule
    \midrule
    \multicolumn{6}{c}{\textbf{Dataset: Wikiplots}}\\
    \midrule
    \textbf{BART}&28.16&11.55&74.81&70.39&90.48\\
    \textbf{\textsc{Hint}}&29.63&12.14&78.83&71.72&91.24\\
    \textbf{\textsc{DiscoDVT}}&28.50&11.68&80.31&73.34&91.94\\
    \midrule
    \textbf{\textsc{Eric}}&\textbf{32.32}&\textbf{13.13}&\textbf{82.45}&\textbf{75.58}&\textbf{92.96}\\
    \midrule
    \midrule
    \multicolumn{6}{c}{\textbf{Dataset: CNN News}}\\
    \midrule
    \textbf{BART}&31.65&14.53&86.95&77.31&92.75\\
    \textbf{\textsc{Hint}}&33.18&15.23&86.17&77.88&93.01\\
    \textbf{\textsc{DiscoDVT}}&32.97&15.08&87.84&77.71&92.85\\
    \midrule
    \textbf{\textsc{Eric}}&\textbf{34.75}&\textbf{15.78}&\textbf{92.96}&\textbf{78.02}&\textbf{93.21}\\
    \bottomrule
    \end{tabular}
    \caption{Automatic evaluation results for different models on the validation sets of Wikiplots and CNN News. We highlight the best performance in \textbf{bold}.}
    \label{tab:val_res}
\end{table}

\begin{table}[!t]
\scriptsize
    \centering
    \begin{tabular}{cc|cc}
\toprule
\multicolumn{2}{c|}{\textbf{Dataset: Wikiplots}}&\multicolumn{2}{c}{\textbf{Dataset: CNN News}}\\
\midrule
\textbf{States}&\textbf{Percentages~(\%)}&\textbf{States}&\textbf{Percentages~(\%)}\\
\midrule
377&3.83&0&1.78\\
425&1.66&3&1.37\\
288&1.52&305&1.37\\
141&1.44&1&1.36\\
68&1.32&151&1.35\\
147&1.25&403&1.34\\
89&1.12&236&1.27\\
73&1.10&478&1.20\\
3&1.02&46&1.18\\
429&1.01&325&1.14\\
\bottomrule
    \end{tabular}
    \caption{Top ten most frequently used states and corresponding percentages in the training sets.}
    \label{tab:code_use}
\end{table}

\begin{table*}[!t]
\scriptsize
    \centering
    \begin{tabular}{p{313pt}p{165pt}}
    \toprule
    \textbf{Text}&\textbf{Example Sentences}\\
    \midrule
    \textit{The 20th century's industrialization leaves the world overcrowded, polluted and suffering global warming due to ``the greenhouse effect''.}
     In 2022, with 40 million people in New York City, housing is dilapidated; homeless people fill the streets; many are unemployed, the few ``lucky'' ones with jobs are only barely scraping by, and food and working technology is scarce. Most of the population survives on rations produced by $\langle\rm e0\rangle$, whose newest product is $\langle\rm e1\rangle$, a green wafer advertised to contain ``high-energy plankton'' from the World Ocean, more nutritious and palatable than its predecessors ``Red'' and ``Yellow'', but in short supply. \textbf{New York City Police Department detective} $\langle\rm \textbf{e2}\rangle_{\textbf{68}}\to\langle\rm \textbf{e0}\rangle_{\textbf{318}}$ \textbf{lives with aged friend and ``book'' (a police analyst) Solomon ``Sol'' Roth.}&\textbf{\textit{State 68}:} they meet the powerful Governor, $\langle\rm e0\rangle_{68}$~($\langle\rm e1\rangle$), his daughter $\langle\rm e2\rangle$, his two grandsons: $\langle\rm e3\rangle$ and $\langle\rm e4\rangle$\newline\newline \textbf{\textit{State 318}:} He is admitted to $\langle\rm e0\rangle_{318}$, mainly because of his genetic mosaicism\\
    \bottomrule
    \end{tabular}
    \caption{A case for the entity coherence modeling experiment on the test set of Wikiplots, where \textsc{Eric} can recognize the incoherence issue while other models fail. \textbf{Left:} The {input}~(in \textit{italic}), the unchanged prefix and the perturbed sentence~(in \textbf{bold}). $A\to B$ means that the original entity A is replaced to B with the subscripts indicating their states predicted by \textsc{Eric}. \textbf{Right:} An example sentence for each state selected from the top ten sentences with closest event representations to the corresponding state vector in the training set of Wikiplots.}
    \label{tab:case_modeling}
\end{table*}

\begin{table*}[!t]
\scriptsize
    \centering
    \begin{tabular}{c|p{220pt}p{185pt}}
    \toprule
    \textbf{State Transitions}&\textbf{The First Sentences}&\textbf{The Second Sentences}\\
    \midrule
    \multirow{5}{*}{\textit{State 462}$\to$\textit{State 301}}&However, in the process, she discovers her husband $\langle\rm \textbf{e3}\rangle_{\textbf{462}}$ is already having another affair with a woman named $\langle\rm {e4}\rangle$ and is now knowingly starting a second affair with ``Anonymous''. & After breaking down over this fact, she goes to visit Archie in the sexy outfit she planned to woo $\langle\rm \textbf{e3}\rangle_{\textbf{301}}$ back with,\\
    \cline{2-3}
    &$\langle\rm \textbf{e3}\rangle_{\textbf{462}}$ realize that once $\langle\rm {e0}\rangle$ gets married it will be just the three of them. & At night, $\langle\rm \textbf{e3}\rangle_{\textbf{301}}$ admits that she is also leaving the group.\\
    \midrule
    \multirow{3}{*}{\textit{State 60}$\to$\textit{State 116}}&She tries to let go of her tomboyish ways to take over $\langle\rm\textbf{e1}\rangle_{\textbf{60}}$'s glamorous lifestyle. & Although she is now dealing with $\langle\rm\textbf{e1}\rangle_{\textbf{116}}$'s job, friends\\
    \cline{2-3}
    &$\langle\rm{e2}\rangle$ spends all her time seeing that $\langle\rm\textbf{e0}\rangle_{\textbf{60}}$ has everything he needs. & $\langle\rm\textbf{e0}\rangle_{\textbf{116}}$ is embarrassed to be seen with his wife\\
    \bottomrule
    \end{tabular}
    \caption{Several examples from the test set of Wikiplots, where an entity~(in \textbf{bold}) transits from a certain state~(in {the first sentence}) to another~(in {the second sentence}). The states are decided by looking up the closest state vectors to the corresponding entity-aware event representations.}
    \label{tab:case_transist}
\end{table*}
\begin{figure*}[!t]
  \centering
  \begin{mdframed}
\includegraphics[width=\linewidth]{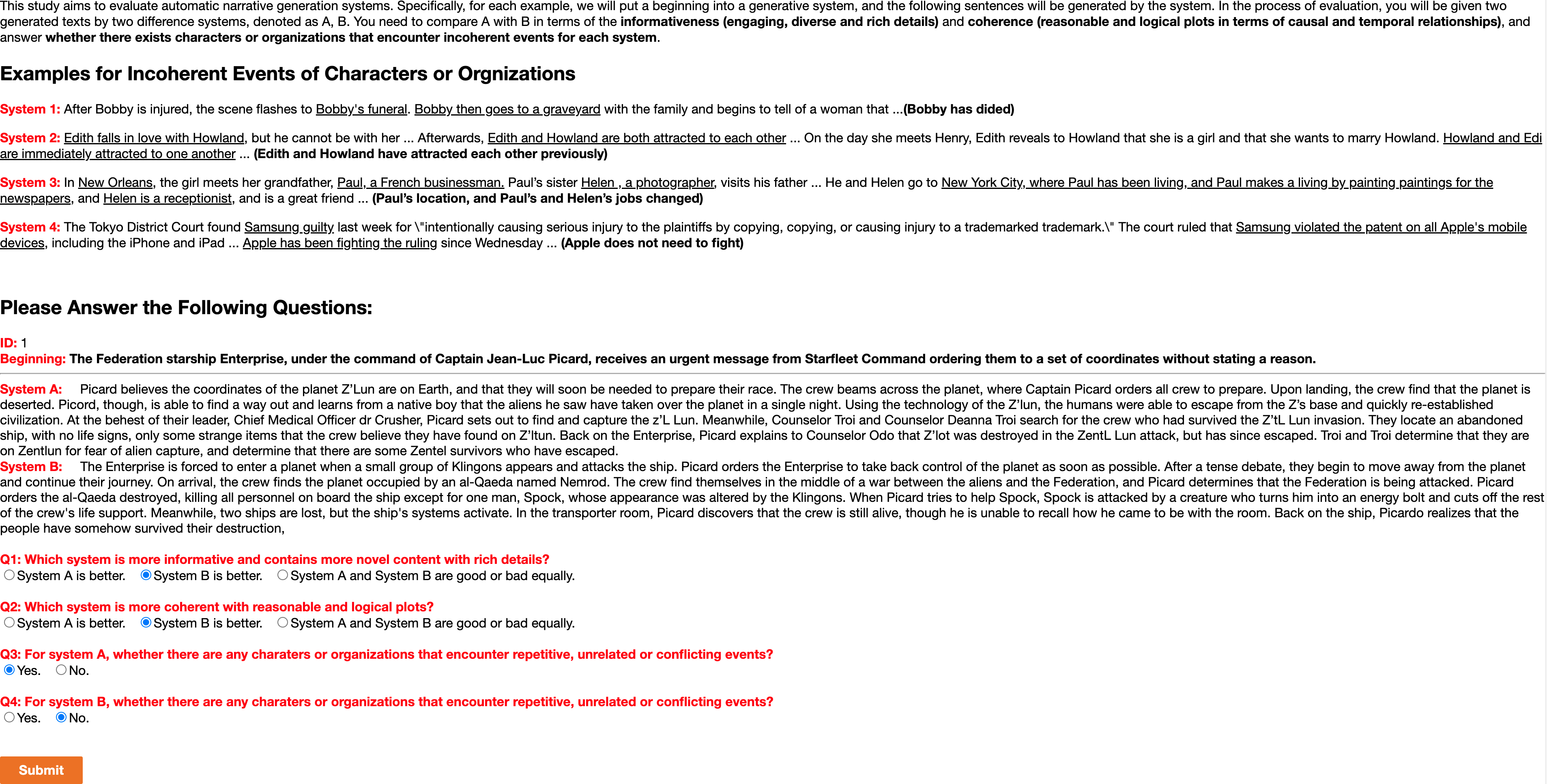}
\end{mdframed}
  \caption{A screenshot of the annotation on AMT for
manual evaluation.}
  \label{fig:man_instruct}
\end{figure*}
\begin{table*}[!t]
\scriptsize
\centering
\begin{tabular}{p{45pt}p{430pt}}
\toprule
\textbf{Input}&The Federation starship Enterprise, under the command of Captain Jean-Luc Picard, receives an urgent message from Starfleet Command ordering them to a set of coordinates without stating a reason.\\
\midrule
\textbf{Truth}&  While en route, they are contacted by Admiral Gromek who informs them that the Enterprise is to rendezvous with a Federation emissary who will brief them on their mission. She refuses to disclose any details, only that Starfleet considers the mission a ``top security matter''. Data reports that the emissary is being transported in a class 8 probe, an unconventional mode of travel and barely large enough to contain a person, traveling at warp 9. Picard notes the evident urgency of the mission. The Enterprise intercepts the probe and beams it aboard, and its passenger is revealed to be a half-Klingon half-human woman named K'Ehleyr. It is immediately clear that K'Ehleyr and lt Worf know each other, and Worf is not pleased to see her. K'Ehleyr informs the command staff that Starfleet has detected a Klingon battlecruiser called the T'Ong, which was launched from the Klingon homeworld over 75 years ago, when the Klingons and the Federation were still at war. The crew has been in suspended animation and are about to awaken, at which point it is feared they will immediately attack the nearest Federation outpost, several of which are nearby and would not be able to adequately defend themselves. As the nearest Klingon ship is three days away, the Enterprise is to intercept them instead. Though K'Ehleyr strongly believes that any attempt to reason with the Klingons will fail,\\
\midrule
\midrule
\textbf{BART}&The Enterprise, after the Enterprise crew leaves, has their first mission briefing to prepare for, however, the crew finds a small force field and a shuttlepod, and are forced to abandon it. Later, \textbf{their commanding officer, Chief Medical Officer dr McCoy}, determines that they are suffering from an unspecified terminal disease and orders the ship repaired. The remaining two \textbf{remaining survivors, dr McCoy} and Lieutenant Commander Riker, are sent out to collect their medication, \textbf{while Chief Medical Advisor dr McCoy has a discussion} with Counselor Troi and \textbf{lt Uhura} over the issue. Captain Picard, Counselor Spock and \textbf{lmdr McCoy} respond to the Enterprise distress call. Picard orders the team to leave and orders a holodeck. They are then ordered to search for their missing shuttlepod and recover their own. A medical officer, dr Leela, takes over command of the station and explains to CounselorTroi that the Enterprise is on a mission to retrieve the shuttlepod. Before the crew can begin, the shuttle pod is struck by a force field, forcing them to abandon the shuttle before they can reach a communications bridge. \textbf{lt Uhura} orders them to go to the bridge.\\
\midrule
\textbf{\textsc{Hint}}&
\textbf{Picard} believes the coordinates of the planet \textbf{Z’Lun} are on Earth, and that they will soon be needed to prepare their race. The crew beams across the planet, where Captain \textbf{Picard} orders all crew to prepare. Upon landing, the crew find that \textbf{the planet is deserted}. \textbf{Picord}, though, is able to find a way out  and learns from a native boy that the aliens he saw have taken over the planet in a single night. \textbf{Using the technology of the Z'lun}, the humans were able to \textbf{escape from the Z's base} and quickly re-established civilization. At the behest of their leader, Chief Medical Officer dr Crusher, Picard sets out to find and capture the \textbf{z'L Lun}. Meanwhile, \textbf{Counselor Troi and Counselor Deanna Troi} search for the crew who had survived the \textbf{Z'tL Lun invasion}. They locate an abandoned ship, with no life signs, only some strange items that the crew believe they have found on \textbf{Z'ltun}. Back on the Enterprise, Picard explains to Counselor Odo that \textbf{Z'lot was destroyed in the ZentL Lun attack, but has since escaped}. \textbf{Troi and Troi} determine that they are on \textbf{Zentlun} for fear of alien capture, and determine that there are some \textbf{Zentel} survivors who have escaped. \\
\midrule
\textbf{\textsc{DiscoDVT}}& En route, \textbf{they meet a race of Vulcans known as ``the Five''}, who have been living on an uninhabited planet, where they have made contact with humans. \textbf{Picard informs them that the humans have been in contact with a race known as the Five}, and that they have been searching for a way to return to Earth. Captain Kirk and Commander Data transport aboard the Enterprise to investigate the Vulcan ship. They find a Starfleet vessel, and a woman, Leda, sitting aboard with the ship.'' Data notes that Leda's appearance is frightening to Kirk and the others, and insists that her name is ``Barbara'' while she is speaking in the Vulcan tongue. He suggests that \textbf{she is actually a human}, and asks that Picard beam her down to the surface to change her appearance, \textbf{but Leda insists that the two will stay with the Enterprise}. \textbf{The Vulcens find the man, who is then forced to be taken into custody, along with his security detail, while the Five are transported to Sickbay}, where Counselor Troi meets with another member of the Five. Meanwhile, on the Enterprise, a mysterious figure called ``The Black Cat'' arrives and attempts to take over the Enterprise's communications systems, but \textbf{they are all too well before the Black Cat is brought down}. \\
\midrule
\textbf{\textsc{Eric}}&The Enterprise is forced to enter a planet when a small group of \underline{Klingons} appears and attacks the ship. \underline{Picard} orders the Enterprise to take back control of the planet as soon as possible. After a tense debate, they begin to move away from the planet and continue their journey. On arrival, the crew finds the planet occupied by an \underline{al-Qaeda} named \underline{Nemrod}. The crew find themselves in the middle of a war between the aliens and the Federation, and \underline{Picard} determines that the Federation is being attacked. \underline{Picard} orders the \underline{al-Qaeda} destroyed, killing all personnel on board the ship except for one man, Spock, whose appearance was altered by the Klingons. When \underline{Picard} tries to help \underline{Spock}, \underline{Spock} is attacked by a creature who turns him into an energy bolt and cuts off the rest of the crew's life support. Meanwhile, two ships are lost, but the ship's systems activate. In the transporter room, \underline{Picard} discovers that the crew is still alive, though he is unable to recall how he came to be with the room. Back on the ship, \underline{Picardo} realizes that the people have somehow survived their destruction, \\
\bottomrule
\end{tabular} 
\caption{Generated cases on Wikiplots. \textbf{Bold} words indicate improper entities or events in terms of coherence. We \underline{underline} the mentioned entities  in the case generated by \textsc{Eric}, which corresponds to the placeholder tokens in Figure~\ref{fig:case}.}
\label{case_gen_wiki}
\end{table*}

\section{Entity State Analysis}\label{state_ana}
\subsection{Entity Mention Control}
Before generating each sentence, \textsc{Eric} first predicts the following mentioned entity, i.e., $\hat{p}_n$.  We find that 96.6\% and 96.0\% of generated sentences on Wikiplots and CNN News, respectively, that mention the predicted $\hat{p}_n$ or do not mention any entities when $\hat{p}_n$ is $\langle\texttt{none}\rangle$. The statistics show that \textsc{Eric} can control which entity to mention in the following sentence with almost perfect accuracy.

\subsection{Entity State Space Utilization} 
We pre-define 512 states in the discrete entity state space $\mathcal{D}$. Their corresponding state vectors are randomly initialized and optimized jointly with the continuous entity-aware event representations through the contrastive framework. Finally, \textsc{Eric} used 501/512 states from $\mathcal{D}$ on the training sets of Wikiplots/CNN News, respectively, showing that the contrastive framework can effectively utilize the capacity of the latent space without any extra entropy penalization~\cite{ji2021discodvt}. 

Furthermore, Table \ref{tab:code_use} shows the top ten most frequently used states and corresponding percentages. Even the most frequent state comprises less than 4\% of the whole state space on both datasets, further indicating that \textsc{Eric} makes the best of the capacity of $\mathcal{D}$.

\subsection{Entity Coherence Modeling}\label{cohe_model}

Table \ref{tab:case_modeling} shows a case to investigate how tracking entity states helps capture the entity coherence. In the perturbed sentence, the original entity $\langle\rm e2\rangle$ is predicted to be in \textit{state 68}, which corresponds to getting together with somebody~(e.g., ``lived with'', ``meet''). When replacing $\langle\rm e2\rangle$ to $\langle\rm e0\rangle$, \textsc{Eric} predicts that $\langle\rm e0\rangle$ is in the \textit{state 318} based on the prefix. \textit{state 318} means that $\langle\rm e0\rangle$ is more likely to be an organization instead of a detective that can live with others, which disagrees with the perturbed sentences. Therefore, we conclude that modeling entity states can effectively helps \textsc{Eric} capture the high-level coherence between events attached to these entities.

Furthermore, Table~\ref{tab:case_transist} shows several examples for two state transition pairs with the highest transition confidence, which is calculated as the product of the cosine distances between the two state vectors to their respective event representations. For \textit{State 462}$\to$\textit{State 301}, both examples relate to ``the entity's marriage is in trouble and then leaves.'' As for \textit{State 60}$\to$\textit{State 116}, both examples relate to ``the entity develops a relationship with another and something trivial happens to them.'' These examples suggest that \textsc{Eric} can capture some patterns of entity state transitions, which helps model inter-event dependencies and high-level coherence.

\section{Manual Annotation Instruction}\label{instruction}
Figure~\ref{fig:man_instruct} shows a screenshot of the
annotation on AMT. 
We paid \$0.4 for annotating an example on average. 
We did not limit the nationalities of annotators,
and did not ask about any personal privacy or collect personal information of annotators in the annotation processes. 


\section{Case Study}\label{case_study}

Table~\ref{case_gen_wiki} presents several cases generated by \textsc{Eric} and strong baselines on Wikiplots. The baselines fail to maintain long-range coherence with unreasonable entity state transitions. For example, in the case generated by BART, at first ``McCoy'' is the ``Chief Medical Officer,'' but then becomes a ``survivor'' to ``collect medication,'' and meanwhile begins a ``discussion'' with the ``Counselor.'' The case generated by \textsc{Hint} also frequently repeats the same entity such as ``Counselor Troi and Counselor Deanna Troi'' and ``Troi and Troi''.  
By contrast, the text generated by \textsc{Eric} has a globally coherent plot, indicating the benefit of tracking entity states and modeling the dependencies between events and entity states explicitly.

\end{document}